\documentclass[letterpaper]{article} 
\usepackage{aaai24}  
\usepackage{times}  
\usepackage{helvet}  
\usepackage{courier}  
\usepackage[hyphens]{url}  
\usepackage{graphicx} 
\usepackage{natbib}  
\usepackage{caption} 
\usepackage{algorithm}
\usepackage{algorithmic}
\usepackage{amsmath,amsfonts}
\usepackage{array}
\usepackage{subfigure}
\usepackage{textcomp}
\usepackage{url}
\usepackage{verbatim}
\usepackage{graphicx}
\usepackage{tabularx}
\usepackage{multirow}
\usepackage{cite}
\usepackage{booktabs}

\newtheorem{definition}{Definition}
\newtheorem{challenge}{Challenge}

\usepackage{newfloat}
\usepackage{listings}
\DeclareCaptionStyle{ruled}{labelfont=normalfont,labelsep=colon,strut=off} 
\floatstyle{ruled}
\newfloat{listing}{tb}{lst}{}
\floatname{listing}{Listing}
\title{Online Boosting Adaptive Learning under Concept Drift \\ for Multistream Classification}
\author{
    En Yu, Jie Lu\thanks{Corresponding Author}, Bin Zhang, Guangquan Zhang\\
}
\affiliations{
    Decision Systems and e-Service Intelligence Laboratory, Australian Artificial Intelligence Institute (AAII),\\ Faculty of Engineering and Information Technology, University of Technology Sydney, Australia\\
    En.Yu@student.uts.edu.au; \{Jie.Lu, Bin.Zhang, Guangquan.Zhang\}@uts.edu.au
}

\begin{document}

\maketitle

\begin{abstract}
Multistream classification poses significant challenges due to the necessity for rapid adaptation in dynamic streaming processes with concept drift. Despite the growing research outcomes in this area, there has been a notable oversight regarding the temporal dynamic relationships between these streams, leading to the issue of negative transfer arising from irrelevant data.
In this paper, we propose a novel \textbf{O}nline \textbf{B}oosting \textbf{A}daptive \textbf{L}earning (OBAL) method that effectively addresses this limitation by adaptively learning the dynamic correlation among different streams. Specifically, OBAL operates in a dual-phase mechanism, in the first of which we design an \textbf{Ada}ptive \textbf{CO}variate \textbf{S}hift \textbf{A}daptation (AdaCOSA) algorithm to construct an initialized ensemble model using archived data from various source streams, thus mitigating the covariate shift while learning the dynamic correlations via an adaptive re-weighting strategy. During the online process, we employ a Gaussian Mixture Model-based weighting mechanism, which is seamlessly integrated with the acquired correlations via AdaCOSA to effectively handle asynchronous drift. This approach significantly improves the predictive performance and stability of the target stream.
We conduct comprehensive experiments on several synthetic and real-world data streams, encompassing various drifting scenarios and types. The results clearly demonstrate that OBAL achieves remarkable advancements in addressing multistream classification problems by effectively leveraging positive knowledge derived from multiple sources.
\end{abstract}

\begin{figure*}[]
\centering
\includegraphics[width=\textwidth]{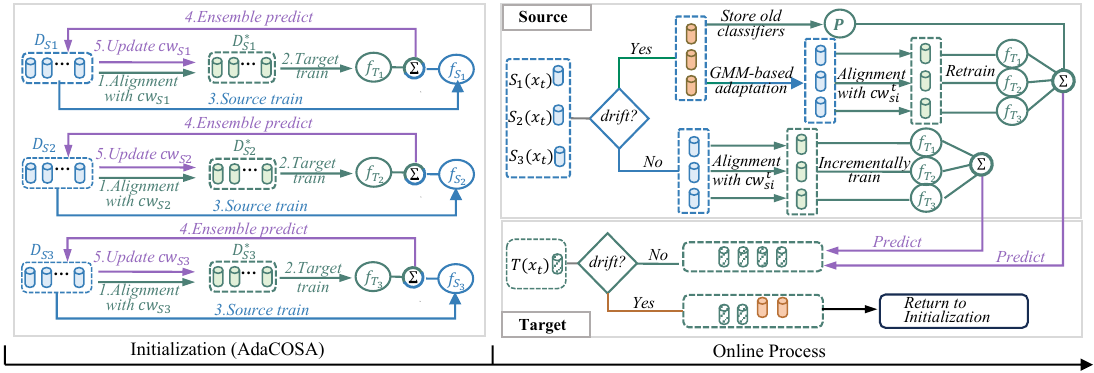}
\caption{Framework of OBAL. The initialization stage is principally devoted to mitigating the problem of covariate shift, along with learning the intricate dynamic correlations that exist between various data streams. In the online phase, the core focus is on the detection and adaptation of asynchronous drift. This stage further integrates the covariate shift alignment and correlation matrices learned during the initial phase, facilitating a seamless ensemble prediction from the source to the target stream.}
\label{overallframework}
\end{figure*}

\section{Introduction}
In various real-world scenarios, such as auto-driving systems, weather forecasts, and industrial production, data is continuously and sequentially generated over time, which is referred to as data streams or streaming data \cite{lu2018learning, zhou2023ods,wang2022learngene}. These data streams are susceptible to changes in their underlying distribution, resulting in concept drift. Consequently, classifiers trained on historical data may fail to predict subsequent samples, leading to a performance decrease \cite{li2022ddg,xu2023open}. Thus, it attracts many researchers to develop efficient learning techniques capable of analyzing streaming data with concept drift in non-stationary environments. To date, prior studies have provided empirical evidence of the efficacy of concept drift adaptation methods in effectively addressing data streams with dynamic distributions. It is worth noting that the majority of existing techniques have been tailored specifically for a single stream with delayed labels \cite{yu2022meta,song2021learning}. However, it is common to encounter scenarios where multiple data streams are generated simultaneously in real-world intelligent systems. For example, data samples continuously stream from sensors in manufacturing systems. These data streams, despite being associated with the same task, often exhibit distinct distributions due to varying data sources \cite{zhou2023multi}. In addition, while data collection is straightforward, the labeling process incurs high time and labor costs, leading to the hybrid multiple streams where massive labeled and unlabeled streams arrive simultaneously \cite{yu2022learn}. 

To tackle this scenario, multistream classification has been proposed, in which a model can be flexibly transferred from labeled source streams to the unlabeled target stream while employing online detection and adaptation working principles. This not only enables the model to adapt to new and unlabeled data streams but also mitigates the expenses and logistical challenges. The multistream classification problem features three major challenges that have to be tackled simultaneously: \textit{1) Scarcity of labels:} this arises from the absence of labels specifically for the target stream, while the source streams possess labeled data; \textit{2) Covariate shift:} this implies that any two data streams exhibit distinct distributions, whether they are different source streams or a source stream and a target stream; and \textit{3) Asynchronous drift:} the source and target streams are susceptible to independent concept drift, which occurs at varying time periods and results in unique effects on the model performance.

In recent years, several approaches have been proposed to address the multistream classification problem by using online domain adaptation and drift handling techniques \cite{chandra2016adaptive,haque2017fusion,pratama2019atl, wang2021evolving}. However, many of these methods have primarily focused on the single-source stream, potentially impeding model performance due to limitations in the quality of the source data. Furthermore, such single-source-based approaches may be prone to overfitting issues. Accordingly, the multi-source configuration is introduced, which enables the acquisition of supplementary information from different source streams, thereby providing more valuable information to build a more accurate and robust model \cite{wang2022self, yang2021concept}. However, leveraging the information from each individual source stream exposes a new challenge: \textit{4) Temporal dynamic correlations} between the source and target streams. In other words, any drift occurring within each stream has the potential to alter the correlation between the source and target streams. It is crucial for the predictive model to adapt promptly, extracting valuable insights from relevant source streams while avoiding the assimilation of irrelevant knowledge from other source streams. 

To address all issues in the multi-stream classification task, we propose the \textbf{O}nline \textbf{B}oosting \textbf{A}daptive \textbf{L}earning (OBAL) method. As shown in Figure \ref{overallframework}, OBAL consists of two stages, the first of which is the initialization phase, where we propose the AdaCOSA algorithm. The fundamental principle of AdaCOSA involves an adaptive interaction between models learned in the original source space and those acquired in the target space, aiming to align the \textit{temporal covariate shift} and explore the \textit{dynamic relationships} between different data streams based on feedback from the target domain. This process reinforces positive knowledge transfer, leading to optimal model migration. 
The second stage involves the online processing phase, during which our primary aim is to detect and adapt to the \textit{asynchronous drift} in each data stream in real-time. To achieve this, we employ the Drift Detection Method (DDM) \cite{gama2004learning} for labeled source streams, as it offers a stable and accurate detection approach. Simultaneously, we utilize the Gaussian Mixture Model (GMM) \cite{oliveira2021tackling} based weighting strategy for asynchronous drift adaptation in these streams. For the unlabeled target stream, we design two sliding windows and continuously monitor their distribution changes to effectively detect drift occurrences. Once a drift is detected in the target stream, it signifies that the dynamic relationships learned in the first stage are no longer applicable, necessitating a return to the first stage for reinitialization. The main contributions of our work can be summarized as follows:

\begin{itemize}
    \item This paper presents a new online ensemble approach (OBAL) for multi-source data stream classification. With the capability to dynamically detect and adapt to concept drift, OBAL demonstrates enhanced effectiveness and stability. Moreover, it offers effortless extensibility in managing diverse data streams.
    \item  A novel algorithm (AdaCOSA) is proposed to align the covariate shift as well as investigate a new dynamic correlation issue between source and target streams. It further enhances positive knowledge transfer and prevents negative transfer effects.
    \item We design a simple yet effective GMM-based module to adapt the asynchronous drift. It orchestrates an ensemble of both historical classifiers and newly trained classifiers on weighted source samples. By accumulating abundant source knowledge, the proposed approach achieves improved prediction accuracy for the target stream.
\end{itemize}

\section{Related Works}
Date stream classification has become an increasingly critical area of research due to the dynamic nature of real-world data streams, i.e., concept drift. Concept drift refers to the underlying data distribution changing over time, which occurs as time $t+1$ if joint distribution $P_{t+1}(X, y) \neq P_t(X, y)$. It poses significant challenges for classifiers to maintain accuracy and adapt promptly. To tackle the concept drift problem, many works have been proposed to ensure the effectiveness and reliability of models \cite{gomes2017adaptive,miyaguchi2019cogra,chiu2020diversity,jothimurugesan2023federated}. 
However, most methods are designed for \emph{single-labeled stream}, which is not suitable for the multi-stream scenario. To fill this research blank, Chandra et al. \cite{chandra2016adaptive} introduce a multi-stream classification framework that utilizes ensemble classifiers for each data stream and incorporates Kernel Mean Matching to reduce the disparity between source and target streams. They further propose the FUSION algorithm \cite{haque2017fusion} to leverage the Kullback Leibler Importance Estimation Procedure for density ratio estimation and covariate shift handling. In addition, some neural-network-based models are proposed to deal with high-dimensional data \cite{yoon2022adaptive}. For example, Autonomous Transfer Learning (ATL) \cite{pratama2019atl} is an online domain adaptation strategy that employs both generative and discriminative phases, combined with Kullback Leibler divergence-based optimization. Moreover, Yu et al. \cite{yu2022meta} propose a meta-learning-based framework to learn the invariant features of drifting data streams and then update the meta model in an online fashion.

In addition, multi-source stream classification is proposed to enhance the robustness by considering the complementary information from different source streams simultaneously. For example, Du et al. \cite{du2019multi} introduced Melanie, which employs a weighted ensemble classifier to transfer knowledge from multiple source streams. It is the first approach capable of simultaneously transferring knowledge from various source streams with concept drift. However, Melanie is a supervised method, which cannot be used for unlabeled data prediction. 

Hence, the AutOmatic Multi-Source Domain Adaptation (AOMSDA) \cite{renchunzi2022automatic} incorporates a central moment discrepancy-based regularizer to leverage the complementary information from multi-source streams, and employs a node weighting strategy to tackle the covariate shift. AOMSDA is a chunk-based method, which means it lacks the ability to dynamically detect the changes in data streams. To address this limitation, Jiao et al. \cite{jiao2022reduced} propose a reduced-space Multistream Classification based on Multi-objective Optimization (MCMO). It seeks a common feature subset to minimize the distribution shift and then uses a GMM to detect and adapt asynchronous drift. However, all these methods determine the correlation between each individual source and target stream as fixed, which does not fully exploit temporal dynamic correlations.

\section{Proposed Method}

\subsection{Problem Definition}

Multi-source-stream classification involves the presence of multiple labeled source streams and one unlabeled target stream. These streams possess interconnected internal representations and share a common label space. The objective of this task is to predict the labels of the target stream by effectively transferring knowledge from the labeled source to the target stream, and it can be defined as follows.
\begin{definition}
    \textbf{Multi-source-stream Classification.} It involves $N$ labeled source streams $S= \left\{S_1, S_2, \cdots, S_N\right\} $ and one unlabeled target stream $T$. Each arrived data sample at time $t$ is represented by $S_i(\boldsymbol{x}_t,y_t)$, where $\boldsymbol{x}_t \in \mathcal{D}^d$ is the $d$-dimensional features, and $y_t$ is the true label of the instance which can only be observed in $S_i$, $i \in \{1,2,\cdots, N\}$. It aims to build a classification model to predict the class label of  $T(\boldsymbol{x}_t)$ using the $S_i(\boldsymbol{x}_t,y_t) $ and $T(\boldsymbol{x}_t)$.
\end{definition}

As mentioned before, four main challenges must be addressed simultaneously in the multistream classification problem, i.e., \textit{scarcity of labels}, \textit{covariate shift}, \textit{asynchronous drift} and \textit{dynamic correlation}. These challenges are defined as follows,

\begin{challenge}
  \textbf{Scarcity of Labels.} This is a major issue in the multistream classification problem. Labeled samples are provided only to the source streams $S_i(\boldsymbol{x}_t, y_t), i \in \{1, 2, \cdots, N\}$, leaving the target stream entirely unlabelled $T(\boldsymbol{x}_t)$. Consequently, the challenge lies in achieving accurate predictions in the target stream, where no labeled samples are available.    
\end{challenge}

\begin{challenge}
     \textbf{Covariate Shift.} Denoting ${P_{S}}_i$ and $P_{T}$ as the distributions from  ${S}_i, i \in {1,2, \cdots, N}$ and ${T}$, all streams at the same time step are related but with covariate shift, i.e., ${P_S}_i(y_t \mid \boldsymbol{x}_t) = {P_S}_j(y_t \mid \boldsymbol{x}_t) =P_T(y_t \mid \boldsymbol{x}_t)$ while ${P_S}_i(\boldsymbol{x}_t) \neq {P_S}_j( \boldsymbol{x}_t)  \neq P_T( \boldsymbol{x}_t)$
\end{challenge}

\begin{algorithm}[t!]
\caption{Initialization (AdaCOSA)}
\label{alg1}
\renewcommand{\algorithmicrequire}{\textbf{Input:}}
\renewcommand{\algorithmicensure}{\textbf{Output:}}
\begin{algorithmic}[1]
    \REQUIRE The archived data batches $D_{Si}, i \in \{1, 2, \cdots N\}$ and $D_T$, Maximum iteration $I_{max}$.
    \ENSURE target classifier $f_{Ti}$, weight vector $\mathbf{cw}_{Si}$.
    \STATE Get the mapped source data $D_{Si}^*$ according to Eq.\ref{mapped_source}.
    \STATE Set up $\beta_n$ and initialize $\mathbf{cw}_{Si}$.
    \FOR{$iter$ = 1 : $I_{max}$}
        \FOR{$i$ = 1 : $N$}
            \STATE Create source classifiers $f_{Si}(x) \leftarrow \{ D_{Si}, Y \}$.
            \STATE Create target classifiers $f_{Ti}(x) \leftarrow \{ D_{Si}^*, Y, \mathbf{cw}_{Si} \}$
            \STATE Predict the instance from $D_{Si}$ using $F_{est}$.
            \STATE Adjust the weight vector $\mathbf{cw}_{Si}$ according to Eq.\ref{update_weight}.
        \ENDFOR       
    \ENDFOR
\end{algorithmic}
\end{algorithm}

\begin{challenge}
    \textbf{Asynchronous Drift.} This refers to the observation of the effect of drift at different times on different independent non-stationary processes that continuously generate data from ${S}= \left\{{S}_1, {S}_2, \cdots, {S}_N\right\}$ and ${T}$. 
     \begin{itemize}
        \item \textbf{Source Drift}: $\exists t$ if ${P_S}_i(\boldsymbol{x}_t) \neq {P_S}_i( \boldsymbol{x}_{t+1}), i \in {1,2, \cdots, N}$ but ${P_T}(\boldsymbol{x}_t) = {P_T}( \boldsymbol{x}_{t+1})$, the drift only occurs in the source stream.
        
        \item \textbf{Target Drift}: $\exists t$ if ${P_S}_i(\boldsymbol{x}_t) = {P_S}_i( \boldsymbol{x}_{t+1}), i \in {1,2, \cdots, N}$ but ${P_T}(\boldsymbol{x}_t) \neq {P_T}( \boldsymbol{x}_{t+1})$, the drift only occurs in the target stream. 
        
        \item \textbf{Concurrent Drifts}: $\exists t$ if ${P_S}_i(\boldsymbol{x}_t) \neq {P_S}_i( \boldsymbol{x}_{t+1}), i \in {1,2, \cdots, N}$ and ${P_T}(\boldsymbol{x}_t) \neq {P_T}( \boldsymbol{x}_{t+1})$, it means drift occurs in both source and target streams.
\end{itemize}
\end{challenge}

\begin{challenge}
    \textbf{Temporal Dynamic Correlation.} The dynamic interplay between source and target streams leads to varying relevance, expressed as ${C}({S}_i(\boldsymbol{x}_t), {T}(\boldsymbol{x}_t))$. At the time $t$, some source streams may possess complementary information ${C}({S}_i(\boldsymbol{x}_t), {T}(\boldsymbol{x}_t)) = {+}$, while others may contain negative information ${C}({S}_i(\boldsymbol{x}_t), {T}(\boldsymbol{x}_t)) = {-}$. The complexity arises as ${C}$ may change over time, such as ${C}({S}_i(\boldsymbol{x}_t), {T}(\boldsymbol{x}_t)) \neq {C}({S}_i(\boldsymbol{x}_{t+\tau}), {T}(\boldsymbol{x}_{t+\tau}))$, disrupting the inherent relationship between the streams.
\end{challenge}
    
To address all challenges, we propose the OBAL method which comprises two stages: initialization (AdaCOSA) and online processing. Next, we will provide a detailed description of these two stages.

\subsection{Adaptive Covariate Shift Adaptation (AdaCOSA)}

To align the covariate shift ${P_S}_i(\boldsymbol{x}_t) \neq {P_S}_j( \boldsymbol{x}_t)  \neq P_T( \boldsymbol{x}_t)$ as well as to explore the temporal dynamic relationship ${C}({S}_i(\boldsymbol{x}_t), {T}(\boldsymbol{x}_t))$ between source and target streams, we propose an AdaCOSA algorithm. Inspired by the CORrelation ALignment (CORAL) method \cite{sun2016return}, the covariance between shifting domains can be aligned by minimizing the distance between the second-order statistics, which provides a stable and effective solution. However, the standard CORAL method is incapable of identifying source instances that are irrelevant to the target, thereby leading to negative transfer effects \cite{wang2019characterizing, yang2021concept}. Furthermore, it fails to address the dynamic relationship between the data streams. As a solution, we propose an adaptive re-weighting strategy to dynamically and iteratively adjust the weights of the source data based on their relevance to the target domain. 

Specifically, given any archived source data batch $D_{Si} = {S}_i(\boldsymbol{X}, Y), i \in \{1, 2, \cdots, N\}$ and the target data batch $D_T = {T}(\boldsymbol{X})$, we first assign a correlation weight vector $\mathbf{cw}_{Si} = [ cw_{Si}^1, cw_{Si}^2, \cdots cw_{Si}^{L_n} ], i \in \{1, 2, \cdots, N\}$ to each source stream, where $L_n$ is the instance number of each archived data batch. Then we can align the shifting covariance by mapping each weighted source data to the target domain using a transformation matrix $A_{Si}$, and the objective can be formulated as,
\begin{equation}
\small
\begin{aligned}
 \min _{A_{Si}}\left\|C_{\hat{S_i}}-C_T\right\|_F^2 
 =\min _{A_{Si}}\left\|A_{Si}^{\top} C_{Si} A_{Si}-C_T\right\|_F^2,
\end{aligned}
\end{equation}
where $\|\cdot\|_F^2$ is the Frobenius norm. $C_{Si}$ and $C_{T}$ are the covariance matrices of $\mathbf{cw}_{Si}D_{Si}$  and $D_{T}$, respectively. $C_{\hat{Si}}$ is the covariance matrix of transformed source features $\mathbf{cw}_{Si} D_{Si}A$, and
\begin{equation}
\small
\begin{aligned}
& C_{S_i}=\operatorname{cov}\left(\mathbf{cw}_{Si} D_{Si}\right)+\operatorname{eye}\left(\operatorname{size}\left(\mathbf{cw}_{Si} D_{Si}, 2\right)\right), \\
& C_T=\operatorname{cov}\left(D_T\right)+\operatorname{eye}\left(\operatorname{size}\left(D_T, 2\right)\right).
\end{aligned}
\label{covariance}
\end{equation}
Then the aligned source data $D_{Si}^*$ can be obtained by the classical whitening and re-coloring strategy \cite{sun2016return} (Please refer to Supplementary S1 for the detailed theoretical analysis),
\begin{equation}
\small
\begin{aligned}
D_{Si}^*= \mathbf{cw}_{Si}  D_{Si}  C_S^{\frac{-1}{2}}  C_T^{\frac{1}{2}}.
\end{aligned}
\label{mapped_source}
\end{equation}

Next, we use a supervised method to train the source classifiers $f_{Si}$ using raw source data $\{ {D}_{Si}, Y\}$. In addition, the covariate-adopted target classifiers $f_{Ti}$ can be learned by using the transformed $\{ D_{Si}^*, Y\}$. 
Finally, we can employ an average ensemble $F_{est}$ that combines models derived from each original source space $f_{Si}$ with those learned in the target space $f_{Ti}$ to re-evaluate the source data iteratively. 

Once the predicted label $\hat{y}_i$ is obtained, it can be used to re-estimate the correlation weights of the source instances because it contains reliable responses from the target domain. In each iteration, if the source instance is predicted mistakenly, this instance may likely conflict with the target stream. Then the effect of this irrelevant data will be diminished in the next iteration by decreasing its training weight.
In contrast, accurate predictions indicate a minimal distance or positive correlation between the source and target domains, resulting in increased training weights to enhance learning. Here, the weight can be updated by,
\begin{equation}
\small
cw_{Si}^t=cw_{Si}^t \cdot e^{-\beta_n\left|\hat{y}_i-y_i\right|},
\label{update_weight}
\end{equation}
where $\beta_n$ is a hyper-parameter defined as $\beta_n = 0.5 \ln \left(1+\sqrt{2 \ln \frac{L_n}{\text { I }_{\max }}}\right)$. $L_n$ is the total number of samples of the archived data batch $D_{S_i}$, and ${I}_{\max}$ is the maximum iterations for adaptive re-weighting.

After several iterations, the instances that exhibit a positive correlation with the target stream will be assigned higher training weights, whereas the training instances that diverge from the target stream will receive lower weights. The detailed process is presented in Algorithm \ref{alg1}. After that, the weight $cw_{Si}$ of each target base classifier can be assigned based on the learned correlation weight and it is calculated by $ cw_{Si}=\frac{1}{L_n} \sum_{t=1}^{L_n} cw_{Si}^t $. Therefore, the final ensemble $f_E$ for the target stream can be formulated as follows:
\begin{equation}
\small
\begin{aligned}
f_E(x)=\frac{cw_{Si}}{\sum_{i=1}^N cw_{Si}} f_{Ti}.
\end{aligned}
\end{equation}

\begin{algorithm}[t!]
\caption{The learning process of OBAL}
\label{alg2}
\renewcommand{\algorithmicrequire}{\textbf{Input:}}
\renewcommand{\algorithmicensure}{\textbf{Output:}}
\begin{algorithmic}[1]
    \REQUIRE Source streams $\left\{{S}_1, {S}_2, \cdots, {S}_N\right\}$, target stream ${T}$, classifier pool $P$, initial sample size $L_n$.
    \ENSURE Predicted labels for target stream.
    \STATE $D_{Si}, D_{T}$  $\leftarrow$  Read first $L_n$ instances from ${S}_i$ and ${T}$.
    \STATE $f_E(x), \mathbf{cw}_{Si}$  $\leftarrow$ initialize according to Algorithm \ref{alg1}.
    \STATE Create $DDM_{Si}$ and $GMM_{Si}$ for source stream.
    \STATE Create $GMM_T$ for target stream.
    \STATE Create detection and reference windows ${W}_{det}$, ${W}_{ref}$. 
    \WHILE{there is incoming data}        
        \FOR{$i$ = 1 : $N$}
        \IF {$DDM_{Si}$ = True}
            \STATE GMM-based adaptation by Eq.\ref{asyn_adaptation}.
            \STATE Weighted alignment and retrain a new classifier.     
        \ELSE
            \STATE Weighted alignment and incrementally update.
        \ENDIF        	
        \ENDFOR	
        \STATE  Move detection window and calculate $\mu_{\text {det}}, \mu_{\text {ref }}$.
        \IF {Eq. \ref{target_drift_con} = True}
            \STATE Remove all base classifiers and return to line 1.
        \ELSE
            \STATE Predict the target sample.
        \ENDIF
    \ENDWHILE
\end{algorithmic}
\end{algorithm}

\begin{table*}[!t]
\centering
\begin{tabular}{lccccccccc}
\toprule
 & SEA & Tree & RBF & Hyperplane & Weather & Kitti & CNNIBN & BBC \\
\midrule
FUSIONs1 & 85.04$_{\pm0.84}$ & 76.98$_{\pm1.11}$ & 82.03$_{\pm1.41}$ & 83.29$_{\pm0.67}$ & 71.04$_{\pm1.50}$ & 54.21$_{\pm2.61}$ & 66.76$_{\pm0.74}$ & 61.76$_{\pm0.09}$ \\

FUSIONs2 & 85.78$_{\pm0.92}$ & 76.74$_{\pm1.00}$ & 83.46$_{\pm1.20}$ & 84.05$_{\pm0.52}$ & 70.65$_{\pm1.32}$ & 52.36$_{\pm2.72}$ & 67.54$_{\pm1.11}$ & 61.26$_{\pm0.43}$ \\

FUSIONs3 & 84.31$_{\pm1.13}$ & 75.21$_{\pm1.07}$ & 81.03$_{\pm1.73}$ & 82.17$_{\pm0.57}$ & 72.17$_{\pm1.17}$ & 50.38$_{\pm2.43}$ & 65.34$_{\pm0.92}$ & 59.86$_{\pm0.19}$ \\

ATLs1 & 88.42$_{\pm1.70}$ & 76.43$_{\pm2.17}$ & 84.53$_{\pm2.01}$ & 86.17$_{\pm1.04}$ & 74.57$_{\pm1.94}$ & 52.78$_{\pm3.78}$ & 62.78$_{\pm1.44}$ & 62.78$_{\pm1.16}$\\

ATLs2 & 88.74$_{\pm1.75}$ & 76.71$_{\pm1.86}$ & 85.21$_{\pm1.85}$ & 87.07$_{\pm1.21}$ & 75.03$_{\pm2.01}$ & 54.01$_{\pm3.09}$ & 65.74$_{\pm1.76}$ & 62.34$_{\pm0.83}$ \\

ATLs3 & 87.62$_{\pm1.01}$ & 76.07$_{\pm2.42}$ & 83.16$_{\pm2.13}$ & 86.01$_{\pm1.49}$ & 74.62$_{\pm1.77}$ & 53.26$_{\pm3.21}$ & 62.65$_{\pm1.38}$ & 60.76$_{\pm0.77}$ \\

Melanie & 89.18$_{\pm0.77}$ & \textbf{78.93}$_{\pm0.61}$ & 86.04$_{\pm0.39}$ & 86.38$_{\pm0.57}$ & 77.74$_{\pm0.89}$ & 50.29$_{\pm1.34}$ & 68.79$_{\pm0.31}$ & \textbf{68.04}$_{\pm0.01}$ \\

AOMSDA & 90.23$_{\pm1.42}$ & 76.87$_{\pm3.47}$ & 85.26$_{\pm2.89}$ & 87.66$_{\pm1.74}$ & 76.55$_{\pm1.41}$ & 67.79$_{\pm3.16}$ & 69.07$_{\pm1.40}$ & 63.36$_{\pm1.07}$ \\

MCMO & 87.46$_{\pm2.12}$ & 77.64$_{\pm1.47}$ & 86.26$_{\pm0.77}$ & 84.04$_{\pm1.42}$ & 76.02$_{\pm3.43}$ & 64.82$_{\pm4.17}$ & 68.83$_{\pm0.89}$ & 60.12$_{\pm1.51}$\\

OBAL (ours) & \textbf{90.98}$_{\pm0.87}$ & 78.45$_{\pm1.01}$ & \textbf{86.78}$_{\pm0.91}$ & \textbf{88.01}$_{\pm1.17}$ & \textbf{79.22}$_{\pm2.07}$ & \textbf{70.29}$_{\pm3.42}$ & \textbf{70.71}$_{\pm0.77}$ & 66.43$_{\pm1.42}$ \\

\bottomrule
\end{tabular}
\caption{Classification accuracy (\%) with the variance of various methods on all benchmarks.}
\label{ACC}
\end{table*}

\subsection{Online Detection and Adaptation}

As stated in Challenge 3, asynchronous concept drifts may occur in either the source or target streams over time. Therefore, for any given stream, it is necessary to continuously monitor its drifting situation in real-time and promptly perform drift adaptation to accommodate the new concept. 

\subsubsection{Source Stream Processing.} For scenarios involving source drift, existing supervised drift detectors such as DDM can be employed, which offers more accurate drift detection because of the leveraging of labels. As a new source sample ${S}_i(\boldsymbol{x}_t)$ arrives, the source classifier predicts its label, and then the drift detector is updated based on the prediction error. If no drift is detected, we will incrementally train the target classifier using the weighted mapped ${S}^*_i(\boldsymbol{x^*}_t, y)$ with its corresponding weight $cw_{si}^t$. Since we have obtained the optimal weights $\mathbf{cw}_{Si} = [ cw_{Si}^1, cw_{Si}^2, \cdots cw_{Si}^{L_n} ]$ for the archived data batch during the initialization stage, we can retrieve the most relevant data from the archived data batch and assign its weights to the new coming data by indexing the minimum $L_2$ distance between new coming and archived data instances. 
However, once a drift is detected within each source stream, an adaptation module should be deployed to handle new concepts. Here, we utilize the GMM to evaluate the distributions of the old and new concepts. GMM assumes several mixture components can model all real-world data, and it is formulated as follows:
\begin{equation}
\small
P(x)=\sum_{k=1}^K P\left(x \mid C_k\right) \cdot w_k,
\label{GMM}
\end{equation}
where $K$ represents the total number of Gaussians or mixture components, and $x$ is the observed multivariate. $w_k$ is a weight that is determined by the observations that constitute $C_k$, and $0 \leq w_k \leq 1, \sum_{k=1}^K w_k=1$. $P\left(x \mid C_k\right)$ represents the likelihood of observation $x$ being assigned to mixture component $C_k$. It can be calculated by using the mean $\mu_k$ and the covariance $\Sigma_k$ of each mixture component $C_k$:
\begin{equation}
\resizebox{0.9\linewidth}{!}{
$P\left(x \mid C_k\right)= 
 \frac{1}{\left(2 \pi^{d / 2} \sqrt{\left|\Sigma_k\right|}\right)} \exp \left(-\frac{1}{2}\left(x-\mu_k\right)^T \Sigma_k^{-1}\left(x-\mu_k\right)\right)$}.
\label{EM}
\end{equation}

According to the Expectation-Maximization (EM) algorithm, all the parameters of different mixture components are randomly initialized using the archived data batch $D_{si}$. Subsequently, it iteratively adjusts the mean and covariance of the mixture component to maximize the likelihood of each mixture component. For a newly incoming instance ${S}_i(\boldsymbol{x}_t)$, its importance weight $aw_{S_i}^{t}$ can be calculated by maximizing the conditional probability of GMM as follows:
\begin{equation}
\small
aw_{S_i}^{t}=\max _{k \in\{1,2, \ldots, K\}} P\left({S}_i(\boldsymbol{x}_t) \mid C_{k}\right).
\label{asyn_adaptation}
\end{equation}

Then, the new coming concept in any source stream can be adapted to the old concept by multiplying $aw_{S_i}^{t}$. Thus, its optimal correlation weight $cw_{si}^t$ with the target stream can also be obtained from the learned $\mathbf{cw}_{Si}$. Finally, a new target base classifier will be created and trained by using weighted mapped ${S}^*_i(\boldsymbol{x^*}_t, y)$ with its corresponding weight $cw_{si}^t$. Note that old base classifiers are no longer trained with new samples but are instead preserved within a base classifier pool denoted as $P$, allowing for their retention. Finally, the joint predictive probability can be ensembled as,
\begin{equation}
\small
\begin{aligned}
    f^E(x)=  \frac{w_{Si}}{\sum_{i=1}^N w_{Si}+\sum_{l=1}^{|P|}  w_{P}} f_{T_i}(x) \\ 
 +\frac{w_{P}}{\sum_{i=1}^N w_{Si}+\sum_{l=1}^{|P|} w_{P}} f_{P}(x),
\end{aligned}
\end{equation}
where $w_{P}$ is the weight of $l$-th classifer in $P$, and $w_{Si}=\frac{1}{n} \sum_{t=1}^{n} aw_{S_i}^{t} * cw_{S_i}^{t} $.


\subsubsection{Target Stream Processing.} 
To detect the drift in the target stream without utilizing labels, we use the archived target data batch $D_{T}$ to initialize a GMM model and deploy two sliding windows to detect the changes over time.
Specifically, we design two sliding windows, i.e., Reference Window $W_{ref}=\left\{T\left(x_{1}\right), \cdots, T\left(x_{n}\right)\}\right.$ and Detect Window $W_{det}=\left\{T\left(x_{n+1}\right), \cdots, T\left(x_{2 n}\right)\}\right.$, where $n$ is the instance number within the window and it is set as $n = L_n$. Then, the average conditional probability of the reference window can be calculated by a point estimation of the mean for the normal distribution,
\begin{equation}
\small
\mu_{\mathrm{ref}}=\frac{1}{n} \sum_{t=1}^n \max _{k \in\{1,2, \ldots, K\}} P\left({T}(\boldsymbol{x}_t) \mid C_k\right).
\end{equation}
The confidence interval estimation of the $\mu_{\mathrm{ref}}$ is known to be $\left[\mu_{\mathrm{ref}}-z_\alpha(\sigma / \sqrt{n}), \mu_{\mathrm{ref}}+z_\alpha(\sigma / \sqrt{n})\right]$, where $\sigma$ is the standard deviation and $z_\alpha$ is the significance level which is set as 3 \cite{kim2017efficient}.
The decision is made that the change has occurred when the point estimation by the mean $\mu_{\mathrm{ref}}$ in the detection window satisfies, 
\begin{equation}
\mu_{\mathrm{det}} \geq \mu_{\mathrm{ref}}+z_\alpha \times \sigma / \sqrt{n}.
\label{target_drift_con}
\end{equation}
Otherwise, $W_{ref}$ and $W_{det}$ move step by step to receive new incoming data, i.e., $W_{ref} = \left\{T\left(x_t\right), \cdots, T\left(x_{n+t-1}\right)\right\}$ and $W_{d e t}=\left\{T\left(x_{n+t}\right), \cdots, T\left(x_{2 n+t-1}\right)\right\}$.
Once a change is detected, the historical base classifier becomes ineffective for classifying target samples. Consequently, all base classifiers are eliminated from the classifier pool, and the model undergoes re-initialization to adapt to the new concepts. The learning process is summarized in Algorithm \ref{alg2}.

\section{Experiments}
In the experiment, we first empirically demonstrated that OBAL consistently outperforms current methods, highlighting both robustness and superiority. Second, we validated the substantial impact of dynamic inter-stream relationships on prediction, emphasizing the effectiveness of the AdaCOSA by ablation study. Additionally, we confirmed OBAL's scalability across various data streams, validating its consistent predictive performance. Finally, we assessed parameter sensitivity, time complexity, and algorithmic cost.


\subsection{Experiment Settings}

\subsubsection{Benchmark Datasets.}
We conduct the experiment on four synthetic datasets (i.e., SEA \cite{street2001streaming}, Tree \cite{liu2020diverse}, RBF \cite{song2021segment}, and Hyperplane \cite{bifet2007learning} ) and four popular real-world datasets (Weather \cite{ditzler2012incremental}, Kitti \cite{geiger2012we}, CNNIBN \cite{vyas2014commercial}, and BBC \cite{vyas2014commercial}), and more detailed descriptions of each dataset and multistream scenario simulation can be found in Supplementary S3 and Table S1.

\subsubsection{Baselines.}
To demonstrate the superiority of our proposed method, we conducted experiments comparing it with five state-of-the-art methods. Among them, the FUSION \cite{haque2017fusion} and ATL \cite{pratama2019atl} algorithms are based on single-source streams, while the Melanie \cite{du2019multi}, AOMSDA \cite{renchunzi2022automatic}, and MCMO \cite{jiao2022reduced} are
specifically designed for the multi-source classification scenario. For FUSION and ATL, we pair each source stream with the target stream, resulting in three distinct groups denoted as \{FUSIONs1, FUSIONs2, and FUSIONs3\} and \{ATLs1, ATLs2, ATLs3\}, respectively.

\subsection{Results Analysis}
\subsubsection{Overall Performance.}
Table \ref{ACC} compares the classification accuracy of OBAL against all baselines on four synthetic and four real-world datasets. Overall, OBAL outperforms all other unsupervised multistream classification methods on both synthetic and real-world datasets, while it performs better than the supervised method (Melanie) on six out of eight datasets. First, compared to single-source-based methods (Fusion and ATL), all multi-source-based methods demonstrate significant improvement. This proves that multiple labeled source streams can provide more discriminative and complementary information, resulting in more accurate and robust predictions. Compared with Melanie, OBAL performs remarkably close to or even surpasses without considering the target labels. 
This is because we not only mitigate the covariate shift but also adaptively adjust sample weights based on the feedback from the target domain. This effectively avoids negative transfer from irrelevant data, thereby ensuring better prediction accuracy. Although AOMSDA and MCMO also consider exploiting the complementary information among multiple source data streams, they ignore the underlying correlation between various streams. In contrast, OBAL employs an adaptive re-weighting approach to iteratively decrease the weights of negative transfer samples and strengthen the weights of positive transfer samples based on the predictive feedback from the target domain. As a result, OBAL achieves the best predictive performance.

\begin{figure}
\centering
\includegraphics[width=\columnwidth]{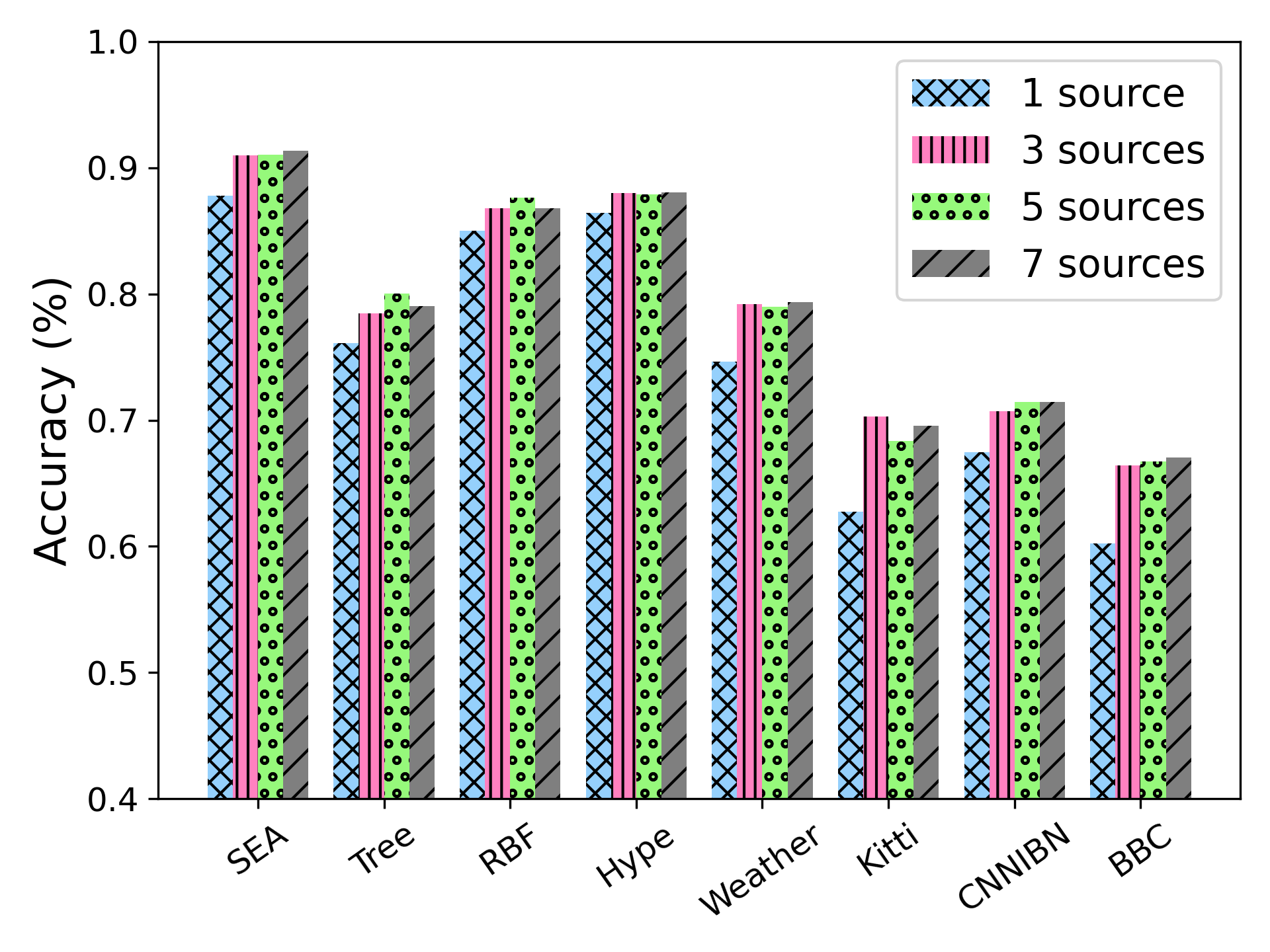}
\caption{The influence of the different number of sources.}
\label{source_numbers}
\end{figure}

\begin{figure*}
\centering
\includegraphics[width=\textwidth]{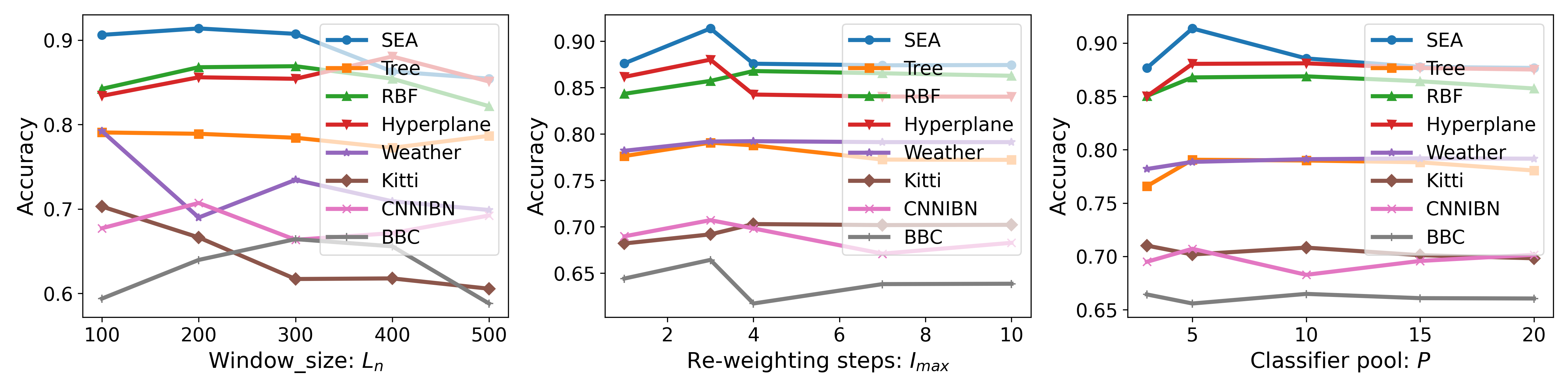}
\caption{The effect of different parameters on classification accuracy.}
\label{parameters}
\end{figure*}
\subsubsection{Ablation Study.}
To validate the rationality of each component and its impact on the overall classification results, we designed three variants of OBAL. As shown in Table \ref{ablation}, OBAL$_{v1}$ as a baseline design does not consider the synchronous drift and covariate shift adaptations. In this situation, each stream is assigned a base classifier and it is updated incrementally. Thus, the performance of OBAL$_{v1}$ is the worst, and significantly lower than that of OBAL on all datasets. This phenomenon highlights the crucial role of concept drift adaptation in dynamic environment learning. OBAL$_{v2}$ considers the synchronous drift in each stream while ignoring the covariate shift alignment.
OBAL$_{v3}$ further employs the traditional CORAL strategy to align the covariate shift which does not explore the dynamic correlation.
By comparing OBAL$_{v2}$ and OBAL$_{v3}$, it can be seen aligning the covariate shift can effectively enhance the performance of the target prediction. Furthermore, the final OBAL highlights the significance of appropriate weights in mitigating the influence of irrelevant source samples and effectively addressing the problem of covariance shift.

\subsubsection{Influence of Source Numbers.}
In this section, we examine the impact of the number of source streams. To ensure a fair comparison with a fixed target stream, we initially sample seven streams from all datasets and vary the number of source streams. Specifically, we evaluate the performance of OBAL using 1, 3, 5, and 7 source streams, respectively. 

This experiment first investigates whether using multi-source streams improves predictive capability compared to a single-source stream. From Figure \ref{source_numbers}, we can observe that the performance of multi-source streams outperforms single-stream performance on all datasets. This indicates that multi-source streams can provide additional complementary information to enhance predictive performance. 
However, as the number of source streams increases, there may be a decline in performance. For example, the performance with five source streams is better than that with seven source streams on the Tree dataset. This may be because as the number of source streams increases, the complexity of the model also increases, which affects its performance. Overall, the performance of OBAL is stable across various sources, which demonstrates that our proposed method can easily adapt to different numbers of data streams.

\subsubsection{Parameter Sensitivity.}
In the proposed OBAL, there are three main parameters affecting the classification performance, including the window size of the initialization stage $L_n$, the re-weighting steps $I_{max}$, and the maximum classifier pool size ${|P|}$. To analyze their impact on the overall performance, we carry out experiments under various values of all parameters on all datasets. Here, we set $L_n \in \{100, 200, 300, 400, 500\}$, $I_{max} \in \{1, 3, 5, 7, 10 \}$ and ${|P|} \in \{1, 5, 10, 15, 20\}$. During the experiment, each parameter is tuned while others are kept fixed, and the various predictive performances are shown in Figure \ref{parameters}. 

Different datasets display varying optimal window sizes due to their unique drift frequencies and periods. For those with frequent drifts, a larger window might encompass multiple concepts, complicating accurate covariate adaptation. Hence, matching the window size to the dataset's drift characteristics is crucial for effective prediction. In the re-weighting phase, the optimal number of iterations for most datasets is three. This is because the algorithm tends to overfit during the initialization phase with an increasing number of iterations. Additionally, as the classifier pool size grows, predictive performance generally improves across datasets, underscoring the importance of retaining historical data. However, after a certain threshold, this performance enhancement plateaus. Detailed parameter settings are shown in Table S2 in the supplementary.

\subsubsection{Time Complexity and Execution Time.} As detailed in Supplementary S4, we analyze the time complexity of OBAL, where the overall complexity is given by $O(L_n log(L_n))O(I_{max} N) + O(L_n) + O(N)O(L_n log(L_n))$. Since $N$ and $I_{max}$ are both quite small, the complexity of OBAL primarily depends on the size of $L_n$. Therefore, we can adjust the value of $L_n$ to execute OBAL efficiently within the available resources. Moreover, Table S3 in the Supplementary compares execution times, revealing that OBAL ranks second after Melanie, underscoring its competitive runtime.  



\begin{table}[!t]
\centering
\fontsize{9pt}{9pt}\selectfont
\begin{tabular}{lccccc}
\toprule
& OBAL$_{v1}$ & OBAL$_{v2}$ & OBAL$_{v3}$ & OBAL  \\
\midrule
SEA & 79.54 & 82.48 & 88.76 & \textbf{90.98} \\
Tree & 72.42 & 74.74 & 77.01 & \textbf{78.45} \\ 
RBF & 79.78 & 81.41 & 84.23 & \textbf{86.78} \\
Hyperplane & 81.42 & 82.35 & 86.34 & \textbf{88.01} \\
Weather & 72.25 & 74.18 & 77.43 & \textbf{79.22} \\ 
Kitti & 62.14 & 64.09 & 68.24 & \textbf{70.29} \\ 
CNNIBN & 63.12 & 67.44 & 69.01 & \textbf{70.71} \\ 
BBC & 58.02 & 62.77 & 64.12 & \textbf{66.43} \\
\bottomrule
\end{tabular}
\caption{Classification accuracy (\%) of OBAL variants.}
\label{ablation}
\end{table}

\section{Conclusion}
In this work, we have addressed a significant gap in multistream classification, where the dynamic relationships between streams have largely been overlooked. This oversight can often result in the issue of negative transfer stemming from irrelevant data. To overcome this challenge, we introduced the Online Boosting Adaptive Learning (OBAL) method, coupled with the proposed AdaCOSA algorithm, effectively exploring the dynamic correlation among various streams. The experiments performed on several synthetic and real-world data streams have shown that our method effectively navigates the dynamic correlations between streams, mitigates covariate shifts, and adeptly handles asynchronous drift using a GMM-based weighting mechanism. The insights gained from this study not only advance the field of multistream classification but also provide a promising direction for future research in adaptive learning across various dynamic data environments.

\section{Acknowledgments}
The work presented in this paper was supported by the Australian Research Council (ARC) under Laureate project FL190100149 and discovery project DP200100700.



\bibliography{aaai24}
\clearpage
\newpage

\appendix

\newtheorem{lemma}{Lemma}
\newtheorem{theorem}{Theorem}
\renewcommand{\thetable}{S\arabic{table}} 
\renewcommand{\thefigure}{S\arabic{figure}}
\renewcommand{\thealgorithm}{S\arabic{algorithm}}
\setcounter{table}{0}
\setcounter{figure}{0}
\setcounter{algorithm}{0}
\section{Supplementary}

\subsection{S1: Theory Analysis}
To derive the solution for Eq.1 presented in this paper, we invoke the subsequent lemma.
\begin{lemma}
    \label{lemma1}
    \cite{cai2010singular} Let $Y$ be a real matrix of rank $r_Y$ and $X$ be a real matrix of rank at most $r$, where $r \leq r_Y$. let $Y=U_Y \Sigma_Y V_Y$ be the SVD of $Y$, and $\Sigma_{Y[1: r]}, U_{Y[1: r]}, V_{Y[1: r]}$ be the largest $r$ singular values and the corresponding left and right singular vectors of $Y$ respectively. Then $X^*=U_{Y[1: r]} \Sigma_{Y[1 ; r]} V_{Y[1 ; r]}^{\top}$ is the optimal solution to the problem of $\min \limits_X\|X-Y\|_F^2$.
\end{lemma}
\begin{theorem}
    \cite{sun2016return} Let $\Sigma^{+}$ be the Moore-Penrose pseudoinverse of $\Sigma$ and $r_{C_T}$ denote the rank of $C_{Si}$ and $C_T$ respectively. Then, $A^*=U_{Si} \Sigma_{Si}^{+\frac{1}{2}} U_{Si}^{\top} {U_{T[1: r]}} {{\Sigma_{T[1: r]}^\frac{1}{2}}} U_{T[1: r]}^{\top}$ is the optimal solution of Eq.1 with $r = min(r_{C_{Si}}, r_{C_T})$.
\end{theorem}

\textbf{\textit{Proof.}} Since $A$ is a linear transformation, $A^{\top} C_{Si} A$ will not increase the rank of $C_{Si}$, i.e., $r_{C_{\hat{S_i}}} \leq r_{C_Si} $. Conducting SVD on $C_{Si}$ and $C_T$, we can get $C_{Si}=U_{Si} \Sigma_{Si} U_{Si}^{\top}$ and $C_T=U_T \Sigma_T U_T^{\top}$, respectively. In order to get the optimal value of $C_{\hat{Si}}$, we consider the following two cases:

\begin{itemize}
    \item \textit{case 1: $r_{C_{Si}}>r_{C_T}$.} The optimal solution is $C_{\hat{Si}}=C_T$. Thus, $C_{\hat{Si}}=U_T \Sigma_T U_T^{\top}=U_{T[1: r]} \Sigma_{T[1: r]} {U_{T[1: r]}}^{\top}$ is the optimal solution of Eq.1 with $r=r_{C_T}$.  
    
    \item \textit{case 2: $r_{C_{Si}} \leq r_{C_T}$.} Then, according to Lemma \ref{lemma1}, $C_{\hat{Si}}=U_T \Sigma_T U_T^{\top}=U_{T[1: r]} \Sigma_{T[1: r]} {U_{T[1: r]}}^{\top}$ is the optimal solution of Eq 1 where $r=r_{C_{Si}}$.
\end{itemize}

\begin{table*}[!t]
\centering
\begin{tabular}{ccccccc}
\hline
 & Datasets & Drift types & Type &\#Instances & \#Features & \#Classes \\ \hline \hline
\multirow{4}{*}{Synthetic} & SEA & Sudden/recurring &Single & 25K * 4 & 3 & 2 \\ 
 & Tree & Sudden/gradual &Single & 5K * 4 & 20 & 2 \\ 
 & RBF & Incremental &Single & 5K * 4& 10 & 2 \\  
 & Hyperplane & Incremental &Single  & 30K* 4 & 4 & 2 \\ \hline \hline
\multirow{4}{*}{Real-world} & Weather & Unknown &Single & 4.5K* 4 & 8 & 2 \\
 & Kitti & Unknown &Single & 6.25K * 4& 55 & 8 \\ 
 & CNNIBN & Unknown &Multistream & 30K * 4& 124 & 2 \\ 
 & BBC & Unknown &Multistream & 30K * 4 & 124 & 2 \\ \hline
\end{tabular}
\caption{Characteristics of all datasets including 3 sources and 1 target stream.}
\label{dataset}
\end{table*}

Therefore, the optimal solution of Eq.1 can be derived as $C_{\hat{Si}}=U_T \Sigma_T U_T^{\top}=U_{T[1: r]} \Sigma_{T[1: r]} {U_{T[1: r]}}^{\top}$ with $r = min(r_{C_{Si}}, r_{C_T})$. Then, to obtain $A$ based on the above analysis, let $C_{\hat{Si}}=A^{\top} C_{Si} A$ and we can get:
\begin{equation*}
A^{\top} C_{Si} A=U_{T[1: r]} \Sigma_{T[1: r]} U_{T[1: r]}^{\top}
\end{equation*}
Since $C_{Si}=U_{Si} \Sigma_{Si} U_{Si}^{\top}$, we have
\begin{equation*}
A^{\top} U_{Si} \Sigma_{Si} U_{Si}^{\top} A=U_{T[1: r]} \Sigma_{T[1: r]} U_{T[1: r]}^{\top}
\end{equation*}
It can be re-written as 
\begin{equation*}
\left(U_{Si}^{\top} A\right)^{\top} \Sigma_{Si}\left(U_{Si}^{\top} A\right)=U_{T[1: r]} \Sigma_{T[1: r]} U_{T[1: r]}^{\top}
\end{equation*}
Assuming $E=\Sigma_{Si}^{+\frac{1}{2}} U_{Si}^{\top} U_{T[1: r]} \Sigma_{T[1: r]}^{\frac{1}{2}} U_{T[1: r]}^{\top}$, then the right side of the above equation can be simplified as $E^{\top} \Sigma_{S}i E$. This gives
\begin{equation*}
\left(U_{Si}^{\top} A\right)^{\top} \Sigma_{Si}\left(U_S^{\top} A\right)=E^{\top} \Sigma_{Si} E
\end{equation*}
Therefore, we can get $U_{Si}^{\top} A$, and the optimal solution of $A$ can be calculated by 
\begin{equation*}
\begin{aligned}
A & =U_{Si} E \\
& =\left(U_{Si} \Sigma_{Si}^{+\frac{1}{2}} U_{Si}^{\top}\right) \left( U_{T[1: r]} {\Sigma_{T[1: r]}^{\frac{1}{2}}  U_{T[1: r]}^{\top}}\right).
\end{aligned}
\end{equation*}

Finally, as analyzed in \cite{sun2016return}, the first part $U_{Si} \Sigma_{Si}^{+\frac{1}{2}} U_{Si}^{\top}$ whitens the source data while the second part $U_{T[1: r]} {\Sigma_{T[1: r]}^{\frac{1}{2}} U_{T[1: r]}^{\top}}$ re-colors it with the target covariance.

\subsection{S2: Alogrithms}
To provide a clearer demonstration of the OBAL's learning process, we present a more detailed algorithmic procedure in Algorithm \ref{alg3}.
\begin{algorithm}[t!]
\caption{The learning process of OBAL}
\label{alg3}
\small
\renewcommand{\algorithmicrequire}{\textbf{Input:}}
\renewcommand{\algorithmicensure}{\textbf{Output:}}
\begin{algorithmic}[1]
    \REQUIRE Multiple labeled source streams $\left\{{S}_1, {S}_2, \cdots, {S}_N\right\}$, Unlabeled target stream ${T}$, classifier pool $P$, initial sample size $L_n$.
    \ENSURE Labels predicted on ${T}$ data.
    \STATE $D_{Si}, D_{T}$  $\leftarrow$  receive $L_n$ instances from ${S}_i$ and ${T}$.
    \STATE $f_E(x), \mathbf{cw}_{Si}$  $\leftarrow$ create iniliazation model via AdaCOSA.
    \FOR{$i$ = 1 : $N$}
        \STATE $DDM_{Si}$ $\leftarrow$ initialize the source drift detector
        \STATE $GMM_{Si}$ $\leftarrow$ create the source Gaussian Mixture Model
    \ENDFOR
    \STATE $GMM_T \leftarrow $: create the target Gaussian Mixture Model. ${W}_{S_i}$.
    \STATE ${W}_{det}$, ${W}_{ref} \leftarrow$: create the detection and reference windows. 
    \WHILE{there is incoming data}
        \STATE \% Source stream processing
    
        \FOR{$i$ = 1 : $N$}
	    \STATE Receive new instances ${S}_i(\boldsymbol{x}_t, y_t)$.
        \IF {$DDM_{Si}$ = True}
            \STATE $P \leftarrow P \cup f_{Ti}$
            \STATE $aw_{S_i}^{t} \leftarrow$: GMM-based adaptation by Eq.8.
            \STATE $cw_{Si}^{t} \leftarrow$: retrieve the correlation weight in $\mathbf{cw}_{Si}$.
            \STATE ${S}^*_i(\boldsymbol{x^*}_t) \leftarrow$: weighted alignment according to Eq.3.     
            \STATE $f_{Ti}$ $\leftarrow$ create new classifier.
            \STATE $DDM_{Si} \leftarrow$: reset.
        \ELSE
            \STATE $DDM_i({S}_i(\boldsymbol{x}_t, y_t)) \leftarrow$: update source detectors.
            \STATE $cw_{Si}^{t} \leftarrow$: retrieve the correlation weight in $\mathbf{cw}_{Si}$.
            \STATE ${S}^*_i(\boldsymbol{x^*}_t) \leftarrow$: weighted alignment according to Eq.3.
            \STATE $f_{Ti} \leftarrow$: incrementally update.
        \ENDIF        	
        \ENDFOR	

        \STATE \% Target stream processing
        \STATE Moving the ${W}_{det}$ 
        \STATE $\mu_{\text {det}}, \mu_{\text {ref }} \leftarrow$: calculate mean conditional probabilities.
        \IF {Eq.11 = True}
            \STATE remove all base classifiers and return to line 1.
        \ELSE
            \STATE $\hat{y}_t \leftarrow$: predict the target sample.
        \ENDIF
    \ENDWHILE
\end{algorithmic}
\end{algorithm}

\begin{table}[!t]
  \centering
  \renewcommand\tabcolsep{12pt} 
    \renewcommand\arraystretch{0.6} 
    \begin{tabular}{lccc}
    \toprule
    \multicolumn{1}{c}{} & $L_n$   &  $I_{max}$ & $|P|$ \\
    \midrule
    SEA   & 200   & 3     & 5 \\
    Tree  & 200   & 3     & 5 \\
    RBF   & 300   & 4     & 10 \\
    Hyperplane & 400   & 3     & 5 \\
    Weather & 100   & 4     & 5 \\
    Kitti & 100   & 5     & 2 \\
    CNNIBN & 200   & 3     & 5 \\
    BBC   & 300   & 3     & 10 \\
    \bottomrule
    \end{tabular}%
  \label{tab:addlabel}%
    \caption{Parameter settings on different datasets.}
\end{table}%

\subsection{S3: Dataset benchmarks}
\label{Appendix_dataset}
\begin{itemize}
    \item SEA \cite{street2001streaming} is a synthetic dataset with two classes consisting of abrupt and recurring drifts. There are three features and the feature's values range from 0 to 10. When $f_1+f_2 \leq \theta$, the data belongs to class 1. Here, $f_1$ and $f_2$ represent the first and second features, respectively. And $\theta$ denotes the threshold for binary classification, which changes from $4 \rightarrow 7 \rightarrow 4 \rightarrow 7$. 
    \item Tree \cite{liu2020diverse} is generated based on a tree structure where features are randomly split, and labels are assigned to the tree leaves. Each attribute is assigned a random value from a uniform distribution to create a new sample, while new concepts are generated by constructing new trees. 
    \item RBF \cite{song2021segment} generator generates data instances using a radial basis function. Centroids are created randomly and assigned a standard deviation value, a weight, and a class label. Incremental drifts are simulated by continuously moving the centroids. 
    \item Hyperplane \cite{bifet2007learning} is also a synthetic dataset based
    on a rotating hyperplane explained in \cite{hulten2001mining}. Positive labels are assigned to examples where $\sum_{j=1}^d \omega_j x_j>\omega_0$, while negative labels are assigned to examples where $\sum_{j=1}^d \omega_j x_j<\omega_0$. Concept drifts can be simulated by adjusting the relative weights.    
    \item Weather \cite{ditzler2012incremental} is a real-world dataset, which pertains to the task of one-step-ahead prediction for determining the occurrence of rainfall. It encompasses weather data spanning a period of 50 years, capturing both the annual seasonal variations and the long-term climate changes.
    \item Kitti \cite{geiger2012we} presents a real-world computer vision challenge that stems from the autonomous driving scenario. The primary objective is to accomplish 3D object detection, employing two high-resolution video cameras—one capturing color images and the other grayscale images—to capture the objects of interest.
    \item TV News Channel Commercial Detection Dataset\footnote{https://archive.ics.uci.edu/dataset/326/tv+news+channel \\ +commercial+detection+dataset}.
 \cite{vyas2014commercial} is a real-world multistream dataset. It comprises of prominent audio-visual features collected from 150 hours of television news broadcasts, including 30 hours each from five news channels (i.e., BBC, CNNIB, CNN, NDTV, and TIMESNOW). All the video shots are recorded in a sequential way and used for commercial or non-commercial detection. In this paper, we designate CNNIBN and BCC as the target streams, while treating the remaining streams as source streams to simulate a multistream classification task. Each individual data stream comprises 30,000 samples, thus providing two substantial benchmarks (CNNIBN and BBC) for analysis and evaluation.

    Specifically, the original dataset is multimodal and contains 5 sets of video features (i.e., video shot length, screen text distribution, motion distribution, frame difference distribution, and edge change ratio) and 7 sets of audio features (i.e., short-term energy, zero crossing rate, spectral centroid, spectral flux, spectral roll-off frequency, fundamental frequency and bag of audio words), totally for 4125 dimensions. In this experiment, we remove the bag of audio words feature and just use the other 11 sets of features. In addition, to retain as much of the original data as possible, we re-sampled all data streams to 30,000 samples.
\end{itemize}

\begin{table*}[htb]
\label{timecost}
\begin{center}
\resizebox{17.5cm}{1.75cm}{
\begin{tabular}{lcccccccccc}
\toprule
 & FUSIONs1 & FUSIONs2 & FUSIONs3 & ATLs1 & ATLs2 & ATLs3 & Melanie  & AOMSDA  & MCMO & OBAL (ours)\\
\midrule
SEA    &50.07&52.45&49.83 &52.31&51.47&50.07 &3.01 &51.30& 43.79 &16.67\\
Tree   &37.76 &36.13 &35.87 &40.42 &39.65 &39.43 &4.17 &37.79 &41.04 &14.93 \\
RBF    &32.57& 33.14& 33.76 &43.01&42.35& 42.50 &3.04 &43.10 &35.04 &16.87\\
Hyperplane &57.41&55.73 &56.20 &52.38&53.09&53.42 &4.10 &54.75 &57.17 &19.37   \\
Weather    &60.41&59.14&57.23 & 52.01&51.49&52.37 &4.97 &51.70 & 89.43 &10.71  \\
Kitti   &71.23 &71.75 &70.93 &94.87 &96.21 &95.43 &10.43 &48.42 & 77.85 &45.11\\
CNNIBN  &409.58&412.37 &408.60 &367.54&359.40 &355.90  &53.15 &354.56 &532.94 & 233.25  \\
BBC      &424.20&407.18 &431.47  &419.87&417.76& 420.39 &53.01 &424.15 &541.02 & 287.93   \\
\bottomrule
\end{tabular}}
\end{center}
\caption{Execution time (s) of Various Methods on All Benchmarks.}
\end{table*}

To simulate the multistream classification scenario, we first sort all samples in descending order according to the probability of each sample $P(x)=\exp \frac{(x-\bar{x})^2}{2 \sigma^2}$ in a Gaussian distribution, which induces the problem of covariate shift. And then the construction of source streams follows a sequential order, with the first source stream being built upon the top $N_i$ samples, followed by the second source stream, the third source stream, and so on up to $(N-1)$-th source stream. The remaining data samples are then assigned to the target stream. All samples selected in each stream will be recovered to the original chronological order to maintain the raw temporal relationship (i.e., Asynchronous drift). Only source streams exclusively consist of labels, whereas the target stream lacks labels, resulting in the scarcity of labels problem. 
     

\begin{figure*}
\centering
\includegraphics[width=\textwidth]{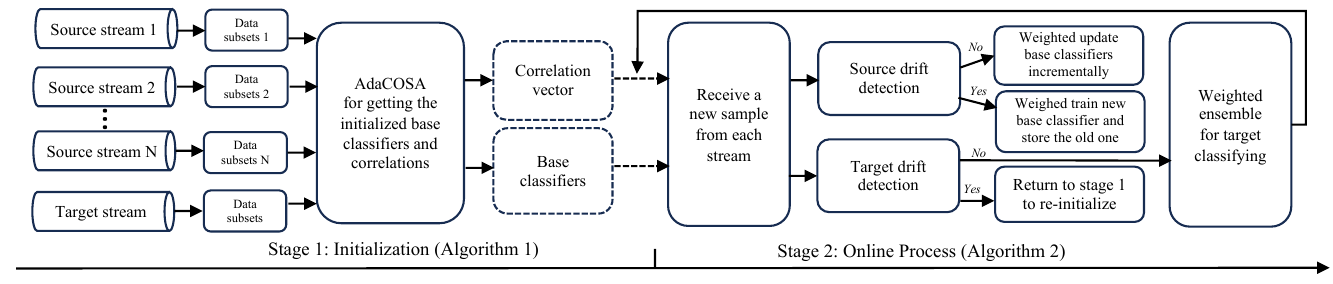}
\caption{High-level illustration of OBAL. The initialization stage is principally devoted to mitigating the problem of covariate shift, along with learning the intricate dynamic correlations that exist between various data streams. In the online phase,  as new source samples arrive, we will incrementally train the base classifiers if no drift is detected. Once a drift is detected within each source stream, a new base classifier will be created and trained. Note that old base classifiers are no longer trained with new samples but are instead preserved within a base classifier allowing for their retention. Furthermore, once the target drift is detected, the historical base classifier becomes ineffective for classifying the target samples. Consequently, all base classifiers are eliminated from the base classifier pool, and the model undergoes re-initialization to adapt to the new concepts.}
\label{overallframework}
\end{figure*}

\subsection{S4: Experiments}
\subsubsection{Platform:} In this study, we implemented the framework using the scikit-multiflow learning library \cite{JMLR:v19:18-251} in Python. All experimental evaluations were conducted on a server equipped with 187GB of memory and powered by an Intel(R) Xeon(R) Gold 6226R CPU @ 2.90GHz.

\subsubsection{Baseline setting:}
Among all compared methods, FUSION\footnote{https://github.com/ahhaque/FUSION}, Melanie\footnote{https://github.com/nino2222/Melanie} and MOMO\footnote{https://github.com/Jesen-BT/MCMO} are ensemble learning-based methods, which are all implemented by Python. To ensure an equitable comparison, we adopt the Hoffeding Tree as the base classifier for all these three methods. On the other hand, ATL\footnote{https://github.com/w870792/ATL} and AOMSDA\footnote{https://github.com/Renchunzi-Xie/AOMSDA} employ the neural network paradigm and are executed in Matlab. In this study, we adhere to the configurations present in their respective original public source codes.

\subsubsection{Time Complexity Analysis}

In our method, we employed the Hoeffding Tree as the base classifier with a time complexity of $O(L_n log(L_n))$. Thus, the time complexities of AdaCOSA is $O(L_n log(L_n))O(I_{max} N)$, where $L_n$, $I_{max}$ and $N$ are the sample number in the initialization stage, re-weighting interaction and source numbers. During the online process, there are three main modules, GMM, DDM, and Hoeffding Tree classifier. We use the EM algorithm to estimate the GMM parameters and the complexity of EM can be regarded as linear $O(L_n)$ in this method. Time complexities of DDM and Hoeffding Tree are $O(L_n)$ and $O(N)O(L_n log(L_n))$. Therefore, the overall complexity of OBAL is $O(L_n log(L_n))O(I_{max} N) + O(L_n) + O(N)(L_n log(L_n))$. In fact, $N$ and $I_{max}$ are quite small, so the complexity of OBAL depends on the size of $L_n$. Therefore, we can tune $L_n$ to execute OBAL within the available resource. 


\begin{table}[!t]
  \centering
    \begin{tabular}{cccc}
    \toprule
    T-test for OBAL & Melanie & AOMSDA & MCMO \\
    p-value &  0.0039     & 0.6045      & 2.6769e$^{-5}$ \\
    \bottomrule
    \end{tabular}%
  \label{tab:addlabel}%
  \caption{Statistical test on SEA dataset under 10 runs}
\end{table}%

\end{document}